\documentclass[runningheads]{llncs}
\usepackage{graphicx}
\usepackage{amsmath,amssymb} % define this before the line numbering.
\usepackage{color}
\usepackage{colortbl}
\usepackage{hyperref}
\renewcommand\UrlFont{\color{red}\rmfamily}

\usepackage{xspace}
\def\eg{\emph{e.g.}\xspace}
\def\ie{\emph{i.e.}\xspace}

\def\etal{\emph{et al.}\xspace}

\usepackage{xcolor}
\begin{document}
\title{T$^2$Net: Synthetic-to-Realistic Translation for Solving Single-Image Depth Estimation Tasks} 
% Replace with your title

\titlerunning{T$^2$Net: Synthetic-to-Realistic Translation for Depth Estimation Tasks}
% Replace with a meaningful short version of your title
%

\author{Chuanxia Zheng\and
	Tat-Jen Cham \and
	Jianfei Cai}
%
%Please write out author names in full in the paper, i.e. full given and family names. 
%If any authors have names that can be parsed into FirstName LastName in multiple ways, please include the correct parsing, in a comment to the volume editors:
%\index{Lastnames, Firstnames}
%(Do not uncomment it, because you may introduce extra index items if you do that, we will use scripts for introducing index entries...)
\authorrunning{Chuanxia Zheng, Tat-Jen Cham and Jianfei Cai}
% Replace with shorter version of the author list. If there are more authors than fits a line, please use A. Author et al.
%

\institute{School of Computer Science and Engineering,\\
	Nanyang Technological University, Singapore\\
	\email{ chuanxia001@e.ntu.edu.sg}, \email{\{astjcham,asjfcai\}ntu.edu.sg}
}

\maketitle              % typeset the header of the contribution
\begin{abstract}

Current methods for single-image depth estimation use training datasets with real image-depth pairs or stereo pairs, which are not easy to acquire. We propose a framework, trained on synthetic image-depth pairs and unpaired real images, that comprises an image translation network for enhancing realism of input images, followed by a depth prediction network. A key idea is having the first network act as a wide-spectrum input translator, taking in either synthetic or real images, and ideally producing minimally modified realistic images. This is done via a reconstruction loss when the training input is real, and GAN loss when synthetic, removing the need for heuristic self-regularization. The second network is trained on a task loss for synthetic image-depth pairs, with extra GAN loss to unify real and synthetic feature distributions. Importantly, the framework can be trained end-to-end, leading to good results, even surpassing early deep-learning methods that use real paired data.
	
\keywords{single-image depth estimation, unpaired images, synthetic data, domain adaptation}
	
\end{abstract}
\section{Introduction} 

Single-image depth estimation is a challenging ill-posed problem for which good progress has been made in recent years, using supervised deep learning techniques \cite{eigen2014depth,eigen2015predicting,liu2016learning,laina2016deeper} that learn the mapping between image features and depth maps from large training datasets comprising image-depth pairs. An obvious limitation, however, is the need for vast amounts of paired training data for each scene type. Building such extensive datasets for specific scene types is a high-effort, high-cost undertaking \cite{saxena2009make3d,silberman2012indoor,Geiger2012CVPR} due to the need for specialized depth-sensing equipment. The limitation is compounded by the difficulty that traditional supervised learning models face in generalizing to new datasets and environments \cite{liu2016learning}.

To mitigate the cost of acquiring large paired datasets, a few unsupervised learning methods \cite{garg2016unsupervised,godard2017unsupervised,kuznietsov2017semi} have been proposed, focused on estimating accurate disparity maps from easier-to-obtain binocular stereo images. Nonetheless, stereo imagery are still not as readily available as individual images, and systems trained on one dataset will find difficulty in generalizing well to other datasets (observed in \cite{godard2017unsupervised}), unless camera parameters and rigs are identical in the datasets.

\begin{figure}[tb!]
	\centering
	\includegraphics[width=\textwidth,clip,trim=0 0mm 0 5mm]{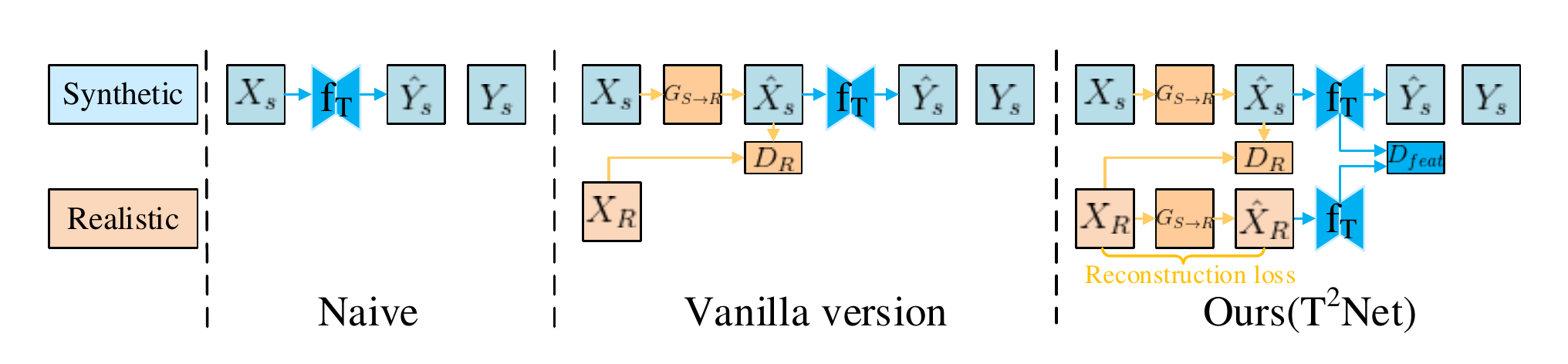}
	
	\caption{Possible approaches to depth estimation using synthetic image-depth pairs $(x_s, y_s)$ and unpaired real images $x_r$. See main text for details.} 
	\label{fig:structure}
\end{figure}

A recent trend that has emerged from the challenge of real data acquisition is the approach of training on synthetic data for use on real data\cite{qiu2016unrealcv,shrivastava2017learning,hoffman2017cycada}, particularly for scenarios in which synthetic data can be easily generated. Inspired by these methods, we have researched a single-image depth estimation method that utilizes synthetic image-depth pairs instead of real paired data, but which also exploits the wide availability of unpaired real images. In short, our scenario is thus: we have a large set of real imagery, but these do not have corresponding ground-truth depth maps. We also have access to a large set of synthetic 3D scenes, from which we can render multiple synthetic images from different viewpoints and their corresponding depth maps. The main goal then is to learn a depth map estimator when presented with a real image.
Consider two of the more obvious approaches:
\begin{enumerate}
	\item Train an estimator using only synthetic image and depth maps, and hope that the estimator applies well to real imagery ({\bf Naive} in fig. \ref{fig:structure}). 
	\item Use a two-stage framework in which synthetic imagery is first translated into the real-image domain using a GAN, and then train the estimator as before ({\bf Vanilla version} in fig.~\ref{fig:structure}).
\end{enumerate}
The problem with 1) is that it is unlikely the estimator is oblivious to the differences between synthetic and real imagery. In 2), while a GAN may encourage synthetic images to map to the distribution of real images, it does not explicitly require the translated realistic image to have any physically-correct relationship to its corresponding depth map, meaning that the learned estimator will not apply well to actual real input. This may be somewhat mediated by introducing some regularization loss to try and keep the translated image ``similar'' in content to the original synthetic image (as in SimGAN\cite{shrivastava2017learning}), but we cannot identify any principled regularization loss functions, only heuristic ones.

In this work, we introduce an interesting perspective on the approach of 2). We propose to have the entire inference pipeline be agnostic as to whether the input image is real or synthetic, \ie it should work equally well regardless. To do so, we want the synthetic-to-realistic translation network to also behave as an identity transform when presented with real images, which is effected by including a reconstruction loss when training with real images.

The broad idea here is that, in a whole spectrum of synthetic images with differing levels of realism, \emph{the network should modify a realistic image less than a more obviously synthetic image}. This is not true of original GANs, which may transform a realistic image into a different realistic image. In short, for the synthetic-to-real translation portion, real training images are challenged with a reconstruction loss, while synthetic images are challenged with a GAN-based adversarial loss \cite{goodfellow2014generative}. This real-synthetic agnosticism is the principled formulation that allows us to dispense with an ad hoc regularization loss for synthetic imagery. When coupled with a task loss for the image-to-depth estimation portion, it leads to an end-to-end trainable pipeline that works well, and does not require the use of any real image-depth pairs nor stereo pairs ({\bf Ours(T$^2$Net)} in fig.~\ref{fig:structure}).

In summary, the main contributions of this work are as follows:
\begin{enumerate}
	\item A novel, end-to-end trainable architecture that jointly learns a synthetic-to-realistic translation network and a task network for single-image depth estimation, without real image-depth pairs or stereo pairs for training. 
	
	\item The concept of a wide-spectrum input translation network, trained by incorporating adversarial loss for synthetic training input and reconstruction loss for real training images, which is justified in a principled manner and leads to more robust translation.
	
	\item The qualitative and quantitative results show that the proposed framework performs substantially better than approaches using only synthetic data, and can even outperform earlier deep learning techniques that were trained on real image-depth pairs or stereo pairs.
\end{enumerate}

\section{Related Work}

For this paper, the two related sets of work are single image depth estimation methods, and unpaired image-to-image translation approaches. 

After classical learning techniques were earlier applied to single-image depth estimation \cite{Hoiem2005,Saxena2008,saxena2009make3d,karsch2012depth,ladicky2014pulling}, deep learning approaches took hold. In \cite{eigen2014depth} a two-scale CNN architecture was proposed to learn the depth map from raw pixel values. This was followed by several CNN-based methods, which included combining deep CNN with continuous CRFs for estimating depth values \cite{liu2016learning}, simultaneously predicting semantic labels and depth maps \cite{wang2015towards}, and treating the depth estimation as a classification task \cite{cao2017estimating}. One common drawback of these methods is that they rely on large quantities of paired images and depths in various scenes for training. Unlike RGB images, real RGB-depth pairs are much scarcer. 

To overcome the above-mentioned problems, some unsupervised and semi-supervised learning methods have recently been proposed that do not require image-depth pairs during training.
In \cite{garg2016unsupervised}, the autoencoder network structure is translated to predict depths by minimizing the image reconstruction loss of image stereo pairs. More recently, this approach has been extended in \cite{godard2017unsupervised,kuznietsov2017semi}, where left-right consistency was used to ensure both good quality image reconstruction and depth estimation. While the data availability for these cases was perhaps not as challenging since special capture devices were not needed, nevertheless they depend on the availability or collection of stereo pairs with highly accurate rigs for consistent camera baselines and relative poses. This dependency makes it particularly difficult to cross datasets (\ie training on one dataset and testing on another), as evidenced by the results presented in \cite{godard2017unsupervised}. To alleviate this problem, an unsupervised adaption method \cite{tonioni2017unsupervised} was proposed to fine-tune a stereo network to a different dataset from which it was pre-trained on. This was achieved by running conventional stereo algorithms and confidence measures on the new dataset, but on much fewer images and at sparser locations.

Separately, several other works have explored image-to-image translation without using paired data. The earlier style-translation networks \cite{gatys2016image,johnson2016perceptual} would synthesize a new image by combining the "content" of one image with the "style" of another image. 
In \cite{liu2016coupled}, the weight-sharing strategy was introduced to learn a joint representation across domains. This framework was extended in \cite{liu2017unsupervised} by integrating variational autoencoders and generative adversarial networks. Other concurrent works \cite{zhu2017unpaired,kim2017learning,yi2017dualgan} utilized cycle consistency to encourage a more meaningful translation. However, these methods were focused on generating visually pleasing images, whereas for us image translation is an intermediate goal, with the primary objective being depth estimation, and thus the fidelity of 3D shape semantics in the translation has overriding importance.

In \cite{shrivastava2017learning}, a SimGAN was proposed to render realistic images from synthetic images for gaze estimation as well as human hand pose estimation. A self-regularization loss is used to force the generated target images to be similar to the original source images. However, we consider this loss to be somewhat ad hoc and runs counter to the translation effort; it may work well in small domain shifts, but is too limiting for our problem. As such, we use a more principled reconstruction loss as detailed in the next sections. More recently, a cycle-consistent adversarial domain adaption method was proposed \cite{hoffman2017cycada} to generate target domain training images for digit classification and semantic segmentation. However this method is too complex for end-to-end training, which we consider to be an important requirement to achieve good results.

%------------------------------------------------------------------------
\section{Method}

Our main goal is to train an image-to-depth network $f_T$, such that when presented with a single RGB image, it predicts the corresponding depth map accurately.

In terms of data availability for training, we assume that we have access to a collection of individual real-world images $x_r$, \emph{without} stereo pairing nor corresponding ground truth depth maps. Instead, we assume that we have access to a collection of synthetic 3D models, from which it is possible to render numerous synthetic images and corresponding depth maps, denoted in pairs of $(x_s, y_s)$.

Instead of directly training $f_T$ on the synthetic $(x_s, y_s)$ data, we expect that the synthetic images are insufficiently similar to the real images, to require a prior image translation network $G_{S\to R}$ for domain adaptation to make the synthetic images more realistic.  However, as discussed previously, existing image translation methods do not adequately preserve the geometric content for accurate depth prediction, or require heuristic regularization loss functions.

Our \emph{key novel insight} is this: instead of training $G_{S\to R}$ to be a narrow-spectrum translation network that translates one specific domain to another, we will train it as a \emph{wide-spectrum} translation network, to which we can feed a range of input domains, \ie synthetic imagery as well as actual real images. The intention is to have $G_{S\to R}$ implicitly learn to apply the minimum change needed to make an image realistic, and consider this the most principled way to regularize a network for preserving shape semantics needed for depth prediction.

\begin{figure}[tb!]
	\centering
	\includegraphics[width=\textwidth]{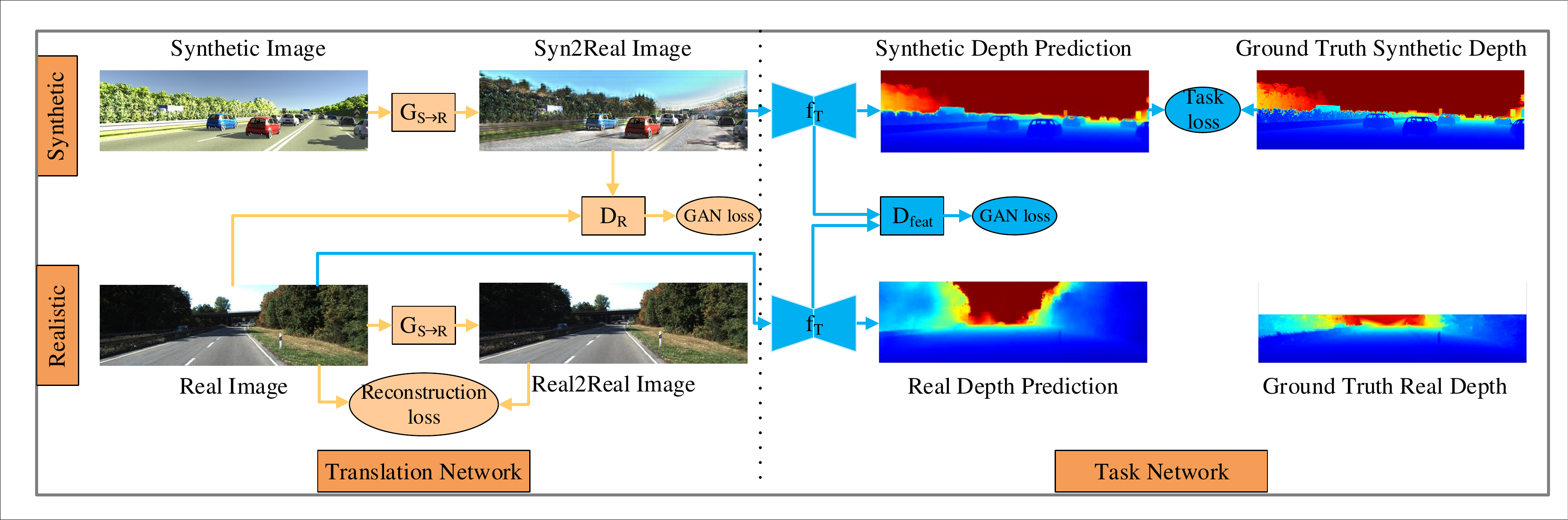}
	\caption{The proposed $T^2$Net consists of the Translation part (left, orange) and Task prediction part (right, blue). See the main text for details.}
	\label{fig:framework}
\end{figure}

To achieve this, we propose the twin pipeline training framework shown in fig.~\ref{fig:framework}, which we call T$^2$Net to highlight the combination of an image \emph{\textbf{t}}ranslation network and a \emph{\textbf{t}}ask prediction network. The upper portion shows the training pipeline with synthetic $(x_s, y_s)$ pairs, while the lower portion shows the training pipeline with real images $x_r$. Note that both pipelines share identical weights for the $G_{S\to R}$ network, and likewise for the $f_T$ network. More specifically:
\begin{itemize}
	\item For real images, we want $G_{S\to R}$ to behave as an autoencoder and apply minimal change to the images, and thus use a \emph{reconstruction loss}.
	\item For synthetic data, we want $G_{S\to R}$ to translate synthetic images into the real-image domain, and use a \emph{GAN loss} via discriminator $D_R$ on the output. The translated images are next passed through $f_T$ for depth prediction, and then compared to the synthetic ground truth depths $y_s$ via a \emph{task loss}.
	\item In addition, we also propose that the inner feature representations of $f_T$ should share similar distributions for both real and translated images, which can be implemented through a feature-based GAN via $D_{\text{feat}}$. 
\end{itemize}
Note that one key benefit of this framework is that it can and should be trained end-to-end, with the weights of $G_{S\to R}$ and $f_T$ simultaneously optimized.

\subsection{Adversarial Loss with Target-Domain Reconstruction}

Intuitively, the gap between synthetic and realistic imagery comes from low-level differences such as color and texture (\eg of trees, roads), rather than high-level geometric and semantic differences. To bridge this gap between the two domains, an ideal translator network, for use within an image-to-depth framework, needs to output images that are impossible to be distinguished from real images and yet retain the original scene geometry present in the synthetic input images. The distribution of real world images can be replicated using adversarial learning, where a generator $G_{S\to R}$ tries to transform a synthetic image $x_s$ to be indistinguishable from real images of $x_r$, while a discriminator $D_R$ aims to differentiate between the generated image $\hat{x}_s$ and real images $x_r$. Following the typical GAN approach \cite{goodfellow2014generative}, we model this minimax game using an \emph{adversarial loss} given by
\begin{equation} \label{eq:1}
\small
\mathcal{L}_{\text{GAN}}(G_{S\to R}, D_R) = \mathbb{E}_{x_r\sim X_R}[\log D_R(x_r)] + \mathbb{E}_{x_s\sim X_S}[\log(1-D_R(G_{S\to R}(x_s)))]
\end{equation}
where generator and discriminator parameters are updated alternately.

However, a vanilla GAN is insufficiently constrained to preserve scene geometry. To regularize this in a principled manner, we want generator $G_{S\to R}$ to behave as a \emph{wide-spectrum} translator, able to take in both real and synthetic imagery, and in both cases produce real imagery. When the input is a real image, we would want the image to remain as much unchanged perceptually, and a \emph{reconstruction loss}
\begin{equation}\label{eq:2}
\mathcal{L}_{r}(G_{S\to R}) = ||G_{S\to R}(x_r) - x_r||_1
\end{equation}
is applied when the input to $G_{S\to R}$ is a real image $x_r$. Note that while this may bear some resemblance to the use of reconstruction losses in CycleGAN \cite{zhu2017unpaired} and $\alpha$-GAN\cite{rosca2017variational}, ours is a unidirectional forward loss, and not a cyclical loss. 

\subsection{Task Loss}
After a synthetic image $x_s$ is translated, we obtain a generated realistic image $\hat{x}_s$, which can still be paired to the corresponding synthetic depth map $y_s$. This paired translated data $(\hat{x}_s, y_s)$ can be used to train the task network $f_T$. Following convention, we directly measure per-pixel difference between the predicted depth map and the synthetic (ground truth) depth map as a task loss:
\begin{equation}\label{eq:3}
\mathcal{L}_{t}(f_T) = ||f_T(\hat{x}_s)-y_s||_1
\end{equation}

We also regularize $f_T$ for real training images. Since real ground truth depth maps are not available during training, a locally smooth loss is introduced to guide a more reasonable depth estimation, in keeping with \cite{heise2013pm,garg2016unsupervised,godard2017unsupervised,kuznietsov2017semi}. As depth discontinuities often occur at object boundaries, we use a robust penalty with an edge-aware term to optimize the depths, similar to \cite{godard2017unsupervised}:
\begin{equation}\label{eq:4}
\mathcal{L}_{s}(f_T) = |\partial_xf_T(x_r)|e^{-|\partial_xx_r|} + |\partial_yf_T(x_r)|e^{-|\partial_yx_r|}
\end{equation}
where $x_r$ is the real world image, and noting that $f_T$ share identical weights in both real and synthetic input pipelines.

In addition, we also want the internal feature representations of real and translated-synthetic images in the encoder-decoder network of $f_T$ to have similar distributions\cite{ganin2015unsupervised}. In theory, the decoder portion of $f_T$ should generate similar prediction results from the two domains when their feature distributions are similar. Thus we further define a feature-level GAN loss as follows:
\begin{equation}\label{eq:5}
\small
\mathcal{L}_{\text{GAN}_f}(f_T, D_{\text{feat}}) =
\mathbb{E}_{f_{\hat{x}_s}\sim f_{\hat{X}_s}}[\log D_{\text{feat}}(f_{\hat{x}_s})] +
\mathbb{E}_{f_{x_r}\sim f_{X_r}}[\log(1-D_{\text{feat}}(f_{x_r}))]
\end{equation}  
where $f_{\hat{x}_s}$ and $f_{x_r}$ are features obtained by the encoder portion of $f_T$ for translated-synthetic images and real images respectively. As noted in \cite{goodfellow2014generative}, the optimal solution measures the Jensen-Shannon divergence between the two distributions.

\subsection{Full Objective}

Taken together, our full objective is:
\begin{align}\label{eq:6}
\mathcal{L}_{\text{T}^2\text{Net}}(G_{S\to R}, f_T, D_R,D_\text{feat}) =
& \mathcal{L}_\text{GAN}(G_{S\to R}, D_R) +  \alpha_{f}\mathcal{L}_{\text{GAN}_{f}}(f_T,D_\text{feat}) \nonumber\\
& + \alpha_{r}\mathcal{L}_{r}(G_{S\to R}) + \alpha_{t}\mathcal{L}_{t}(f_T) + \alpha_{s}\mathcal{L}_{s}(f_T) 
\end{align}
where $\mathcal{L}_\text{GAN}$ encourages translated synthetic images to appear realistic, $\mathcal{L}_{r}$ spurs translated real images to appear identical, $\mathcal{L}_{\text{GAN}_{f}}$ enforces closer internal feature distributions, $\mathcal{L}_t$ promotes accurate depth prediction for synthetic pairs, and $\mathcal{L}_s$ prefers an appropriate local depth variation for real predictions. In our end-to-end training, this objective is used in solving for optimal $f_T$ parameters:
\begin{equation}
f_T^* = \arg \min_{f_T} \min_{G_{S\to R}} \max_{D_R,D_\text{feat}}
\mathcal{L}_{\text{T}^2\text{Net}}(G_{S\to R}, f_T, D_R,D_\text{feat})
\end{equation}
%which can be applied to predict depth maps from test input of real images.

\subsection{Network Architecture}

The transform network, $G_{S\to R}$, is a residual network (ResNet) \cite{he2016deep} similar to SimGAN \cite{shrivastava2017learning}. Limited by memory constraints and the large size of scene images, one down-sampling layer is used in our model and the output is only passed through 6 blocks. For the image discriminator networks, we use PatchGANs \cite{shrivastava2017learning,zhu2017unpaired}, which have produced impressive results by discriminating locally whether image patches are real or fake.

The task prediction network is inspired by \cite{godard2017unsupervised}, which outputs four predicted depth maps of different scales. Instead of encoding input images into very small dimensions to extract global information, we instead use multiple dilation convolutions \cite{YuKoltun2016} with a large feature size to preserve fine-grained details. In addition, we employ different weights for the paths with skip connections \cite{ronneberger2015u}, which can simultaneously process larger-scale semantic information in the scene and yet also predict detailed depth maps. The use of these techniques allows our task prediction network $f_T$ to achieve state-of-the-art performance in our own real-supervised benchmark method (training $f_T$ on pairs of real images and depth), even when the encoder portion of $f_T$ is primarily based on VGG, as opposed to a more typical ResNet50-type network used in other methods \cite{godard2017unsupervised,kuznietsov2017semi}. 

%------------------------------------------------------------------------
\section{Experimental Results}

We evaluated our model on the outdoor KITTI dataset\cite{Geiger2012CVPR} and the indoor NYU Depth v2 dataset \cite{silberman2012indoor}. During the training process, we only used unpaired real images from these datasets in conjunction with synthetic image-depth pairs, obtained via SUNCG \cite{song2017semantic} and vKITTI \cite{gaidon2016virtualworlds} datasets, in our proposed framework.

\subsection{Implementation Details}
\label{sec:implementation}

\paragraph{\bf Training Details:} In order to control the effect of GAN loss, we substituted the vanilla negative log likelihood objective with a least-squares loss \cite{mao2016multi}, which has proven to be more stable during adversarial learning \cite{zhu2017unpaired}. Hence, for GAN loss $\mathcal{L}_\text{GAN}(G_{S\to R}, D_R)$ in (\ref{eq:1}), we trained $G_{S\to R}$ by minimizing
\begin{equation*}
\mathbb{E}_{x_s\sim X_s}[(D_R(G_{S\to R}(x_s))-1)^2]
\end{equation*}
and trained $D_R$ by minimizing
\begin{equation*}
\mathbb{E}_{x_r\sim X_r}[(D_R(x_r)-1)^2] + \mathbb{E}_{x_s\sim X_s}[D_R^2(G_{S\to R}(x_s))].
\end{equation*}
A similar procedure was also applied for the GAN loss in (\ref{eq:5}).

We trained our model using PyTorch. During optimization, the weights of different loss components were set to $\alpha_f$=0.1, $\alpha_r$=40, $\alpha_t$=20, $\alpha_s$=0.01 for indoor scenes and $\alpha_f$=0.1, $\alpha_r$=100, $\alpha_t$=100, $\alpha_s$=0.01  for outdoor scenes. For both indoor and outdoor datasets, we used the Adam solver \cite{kingma2014adam}, setting $\beta_1$=0.5, $\beta_2$=0.9 for the adversarial network and $\beta_1$=0.95, $\beta_2$=0.999 for the task network. All networks were trained from scratch, with a learning rate of $10^{-4}$(task network) and $2\!\times\!10^{-5}$ (translation network) for the first 10 epochs and a linearly decaying rate for the next 10 epochs. In addition, as the indoor synthetic images and real NYUDv2 images are visually quite different, they are easily distinguished by the discriminator. To balance the minimax game, we updated $G_{S\to R}$ five times for each update of $D_R$ during the indoor experiments. Please see the supplementary material and our code\footnote{Available at \UrlFont{https://github.com/lyndonzheng/Synthetic2Realistic}} for more details.

\paragraph{\bf Our $f_T$-only Benchmark Models} Besides our full T$^2$Net model, we also tested our partial model which comprised solely the $f_T$ task prediction network. We evaluated this in two scenarios: (1) an \textbf{``all-real''} scenario, in which we used real image and depth map pairs for training, for which we would expect to \emph{upper bound} our full model performance, and (2) an \textbf{``all-synthetic''} scenario, in which we used only synthetic image-depth pairs and eschewed even unpaired real images, for which we would expect to \emph{lower bound} our full model performance.

\paragraph{\bf Evaluation Metrics:} We evaluated the performance of our approach using the depth evaluation metrics reported in \cite{eigen2014depth}:

\begin{equation*}
\small
\renewcommand{\arraystretch}{1.6}
\setlength{\arraycolsep}{0pt}
\begin{array}{ll}
{\bf RMSE(log):} \sqrt{\frac{1}{|T|}\sum_{i=1}^{T}||\log \hat{y}_{r,i} - \log y_{r,i} ||^2} & {\bf RMSE:} \sqrt{\frac{1}{|T|}\sum_{i=1}^{T}||\hat{y}_{r,i} - y_{r,i}||^2}\\
{\bf Sq.\ relative:} \frac{1}{|T|}\sum_{i=1}^{T}\frac{||\hat{y}_{r,i} - y_{r,i}||^2}{y_{r,i}} & {\bf Abs\ relative:} \frac{1}{|T|}\sum_{i=1}^{T}\frac{|\hat{y}_{r,i} - y_{r,i}|}{y_{r,i}} \\
{\bf Accuracy:\ \% \ of\ y_{r,i}\ s.t.}\ max(\frac{\hat{y}_{r,i}}{y_{r,i}}, \frac{y_{r,i}}{\hat{y}_{r.i}})=\delta < thr
\end{array}
\end{equation*}

\subsection{NYUDv2 Dataset}

\begin{figure}[tb!]
	\centering
	\includegraphics[width=\textwidth]{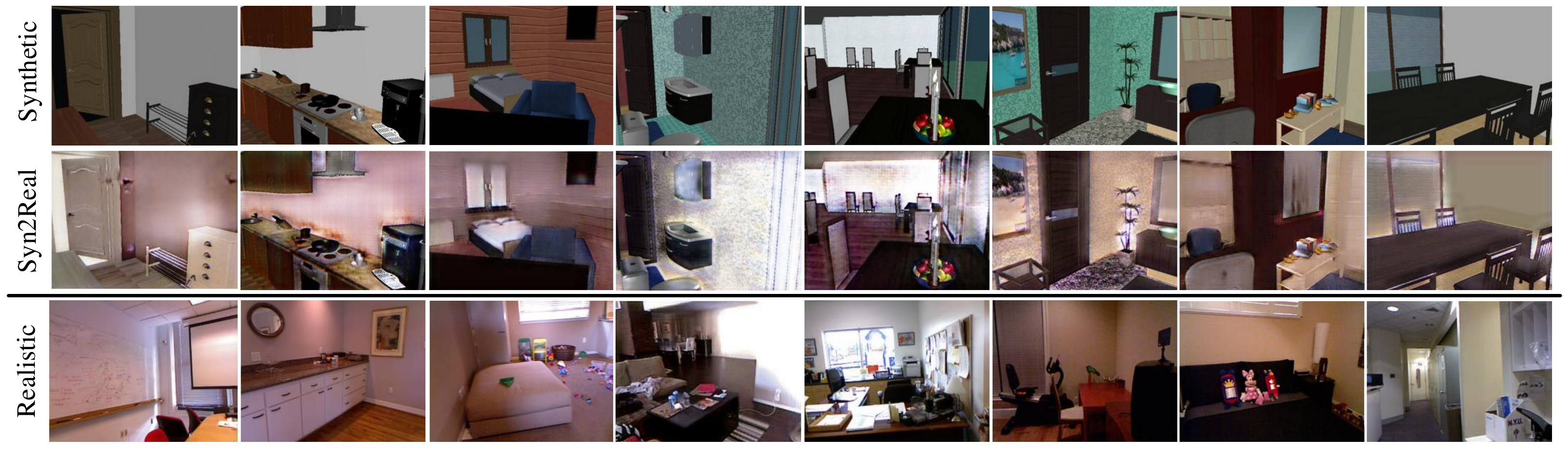}
	\caption{Example output of our translation network for SUNCG\cite{song2017semantic} renderings. Top: synthetic images rendered from SUNCG. Middle: corresponding images after $G_{S\to R}$ translation. Bottom: real images from NYUDv2 \cite{silberman2012indoor} (no correspondence to above rows). 
	}
	\label{fig:syn2real_nyu}
\end{figure}

\paragraph{\bf Synthetic Indoor Dataset:} To generate the paired synthetic training data, we rendered RGB images and depth maps from the SUNCG dataset \cite{song2017semantic}, which contains 45,622 3D houses with various room types. We chose the camera locations, poses and parameters based on the distribution of real NYUDv2 dataset \cite{silberman2012indoor} and retained valid depth maps using the criteria presented in \cite{song2017semantic}: a) valid depth area (depth values in range of 1m to 10m) larger than 70\% of image area, and b) more than two object categories in the scene. In total we generated 130,190 valid views from 4,562 different houses, with samples shown in fig.~\ref{fig:syn2real_nyu}. 

\paragraph{\bf Translated Results:} Fig.~\ref{fig:syn2real_nyu} shows sample output from translation through $G_{S\to R}$. We observe that the visual differences between synthetic and real images are obvious: colors, textures, illumination and shadows in real scenes are more complex than in synthetic ones. Compared to synthetic images, the translated versions are visually more similar to real images in terms of low-level appearance.

\begin{table}[tb!]
	\begin{center}
		\renewcommand{\arraystretch}{1.4}
		\setlength{\arrayrulewidth}{0.8pt}
		\caption{Depth estimation results on NYUDv2 dataset\cite{silberman2012indoor}. \emph{Gray rows indicate methods in which training is conducted \textbf{without} real image-depth pairs. Best supervised results are marked with *, while best unsupervised results are in bold}.}
		\begin{tabular}{|l|c|c|c|c|c|c|c|}
			\hline
			\multicolumn{1}{|c|}{} &  \multicolumn{4}{c|}{\cellcolor[rgb]{0.6,0.8,1.0} lower is better} &  \multicolumn{3}{c|}{\cellcolor[rgb]{0.0,0.8,1.0} higher is better} \\
			\hline
			Method & \cellcolor[rgb]{0.6,0.8,1.0} {\tiny Abs Rel} & \cellcolor[rgb]{0.6,0.8,1.0} {\tiny Sq Rel} & \cellcolor[rgb]{0.6,0.8,1.0} {\tiny RMSE} & \cellcolor[rgb]{0.6,0.8,1.0} {\tiny RMSE log} & \cellcolor[rgb]{0.0,0.8,1.0} {\tiny $\delta$<1.25} & \cellcolor[rgb]{0.0,0.8,1.0} {\tiny $\delta$<1.25$^2$}  & \cellcolor[rgb]{0.0,0.8,1.0} {\tiny $\delta$<1.25$^3$} \\
			\hline
			Ladicky et al. \cite{ladicky2014pulling} & - & - & - & - & 0.542 & 0.829 & 0.940 \\
			Eigen et al.\cite{eigen2014depth} Fine & 0.215 & 0.212 & 0.907 & 0.285 & 0.611 & 0.887 & 0.971 \\
			Liu et al. \cite{liu2016learning} & 0.213 & - & 0.759 & - & 0.650 & 0.906 & 0.976 \\
			Eigen et al.\cite{eigen2015predicting} (VGG) & 0.158 & 0.121$^*$ & 0.641 & 0.214 & 0.769 & 0.950$^*$ & 0.988$^*$ \\
			\hline 
			\rowcolor[rgb]{0.9,0.9,0.9}
			Baseline, train set mean & 0.439  & 0.641 & 1.148 & 0.415 & 0.412 & 0.692 & 0.856 \\  
			\hline
			Our $f_T$, all-real &  0.157$^*$ & 0.125 & 0.556$^*$ & 0.199$^*$ & 0.779$^*$ & 0.943 & 0.983 \\
			\rowcolor[rgb]{0.9,0.9,0.9}
			Our $f_T$, all-synthetic & 0.304 & 0.394 & 1.024 & 0.369 & 0.458 & 0.771 & 0.916 \\
			\rowcolor[rgb]{0.9,0.9,0.9}
			Our T$^2$Net, $D_\text{feat}$ only & 0.320 & 0.405 & 0.991 & 0.343 & 0.480 & 0.792 & 0.933 \\
			\rowcolor[rgb]{0.9,0.9,0.9}
			Our T$^2$Net, $D_\text{image}$ only & 0.274 & 0.336 & 1.001 & 0.325 & 0.496 & 0.814 & 0.938 \\
			\rowcolor[rgb]{0.9,0.9,0.9}
			Our full T$^2$Net & {\bf 0.257} & {\bf 0.281} & {\bf 0.915} & {\bf 0.305}  & {\bf 0.540} & {\bf 0.832} & {\bf 0.948} \\
			\hline
		\end{tabular}
		\label{table:indoor}
	\end{center}
\end{table}

\begin{figure}[tb!]
	\centering
	\includegraphics[width=\textwidth]{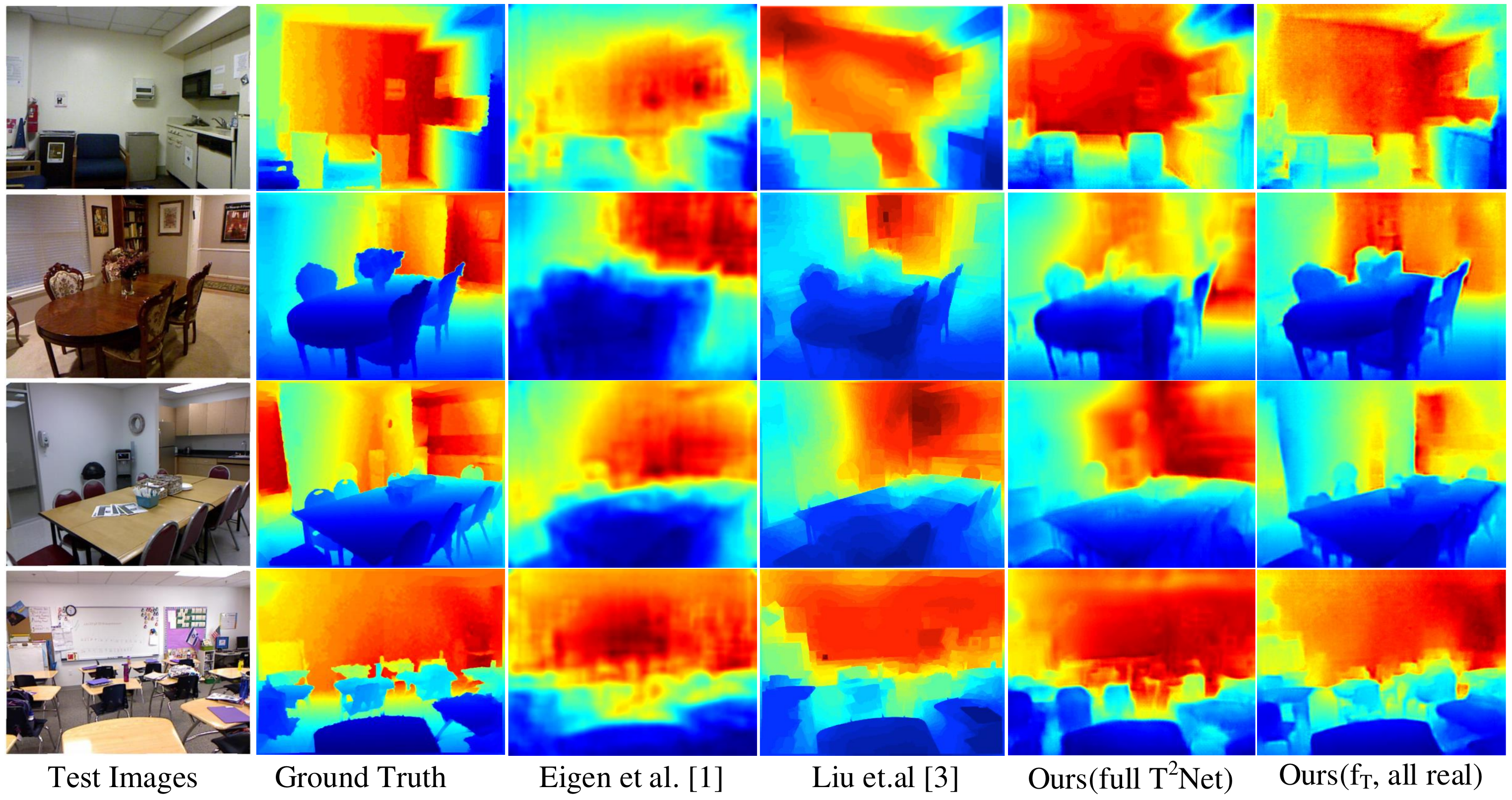}
	\caption{Qualitative results on NYUDv2. All results are shown as relative depth maps (red=far, blue=close). See text for details. }
	\label{fig:depth_indoor}
\end{figure}

\paragraph{\bf Depth Estimation Results:} In table \ref{table:indoor}, we report the performance of our models (varying different application of the two GANs) as compared to latest state-of-the-art methods on the public NYUDv2 dataset. In the indoor dataset, these previous works were all based on supervised learning with real image-depth pairs. The gray rows highlight methods in which real image-depth pairs were \emph{not} used in training. The {\bf train-set-mean} baseline used the mean synthetic depth map in the training dataset as prediction, with the results providing an indication of the correlation between depth maps in the synthetic and real datasets.
We also present results from our $f_T$-only benchmark models in the ``all-real'' and ``all-synthetic'' setups (see section~\ref{sec:implementation}), which we expect to provide the upper bound and lower bound of our model respectively.

Our proposed models produced a clear gap to the train-set-mean baseline and the synthetic-only benchmark. While our models were unable to outperform the latest fully-supervised methods trained on real paired data, the full T$^2$Net model was even able to outperform the earlier supervised learning method of \cite{ladicky2014pulling} on two of the three metrics, despite not using real paired data.

We also show qualitative results in fig.~\ref{fig:depth_indoor}. Although the absolute values of our predicted depths were not as accurate as the latest supervised learning methods, we observe that our T$^2$Net model generates reasonably good relative depths with distinct furniture shapes, even without using real paired training data.

\begin{figure}[tb!]
	\centering
	\includegraphics[width=\textwidth]{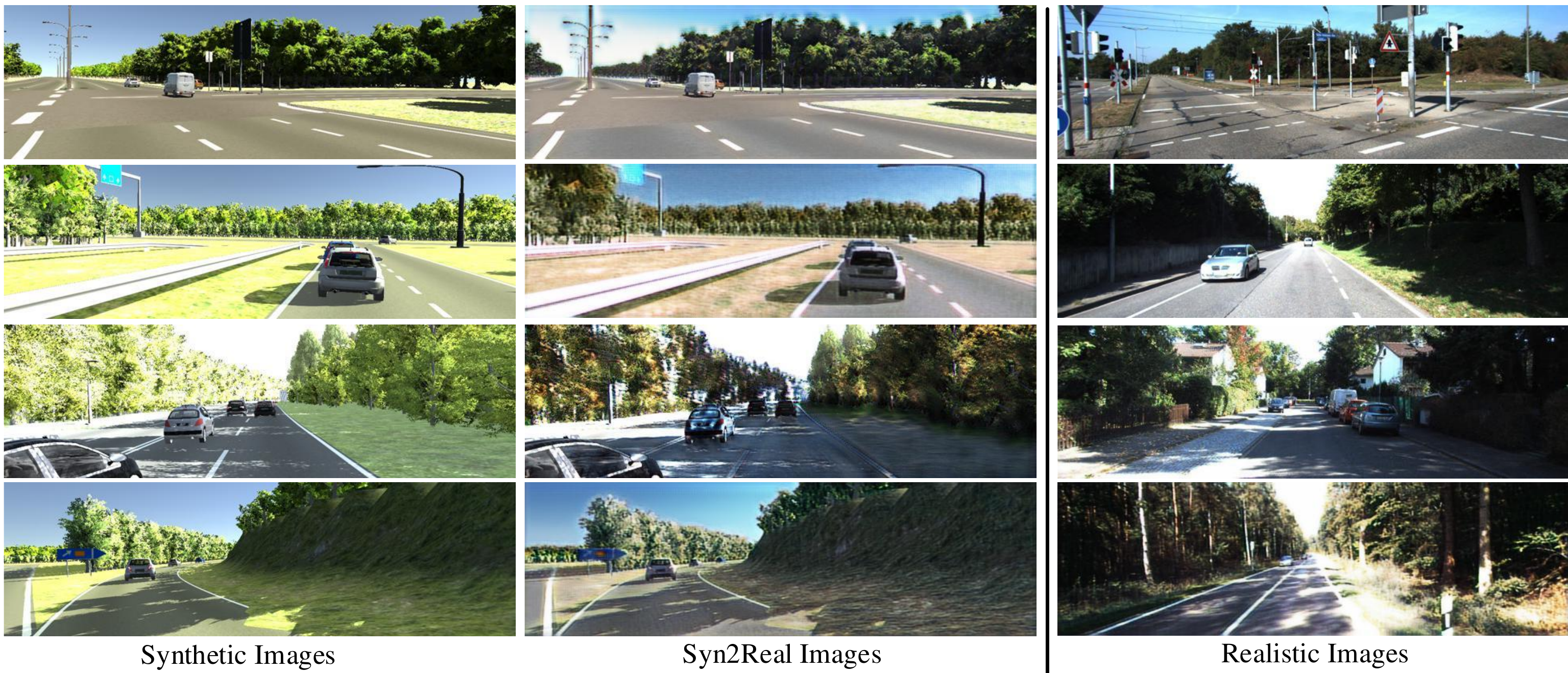}
	\caption{Example translated images for the outdoor vKITTI dataset \cite{gaidon2016virtualworlds}. (Right) the images in real KITTI. (Left) synthetic images from vKITTI and translated images.}
	\label{fig:syn2real_kitti}
\end{figure}

\subsection{KITTI Dataset}

\paragraph{\bf Data Preprocessing:} We used Virtual KITTI (vKITTI) \cite{gaidon2016virtualworlds}, a photo-realistic synthetic dataset that contains 21,260 image-depth paired frames generated from different virtual urban worlds. The scenes and camera viewpoints are similar to the real KITTI dataset \cite{Menze2015CVPR}; see samples in fig.~\ref{fig:syn2real_kitti}. However, the ground truth depths in vKITTI and KITTI are quite different. The maximum sensed depth in a real KITTI image is typically on the order of 80m, whereas vKITTI has precise depths to a maximum of 655.3m. To reduce the effect of ground truth differences, the vKITTI depth maps were clipped to 80m.

\paragraph{\bf Translated Results:} Figure \ref{fig:syn2real_kitti} shows examples of synthetic, translated, and real images from the outdoor datasets. As shown, the translated images have substantially greater resemblance to the real images than the synthetic images. Our translation network can visually replicate the distributions of colors, textures, shadows and other low-level features present in the real images, and meanwhile preserve the scene geometry of the original synthetic images.

\paragraph{\bf Depth Estimation Results:} In order to compare with previous work, we used the test split of 697 images proposed in \cite{eigen2014depth}. Following \cite{godard2017unsupervised}, we chose 22,600 RGB images from the remaining 32 scenes for training the translation network. As before, we did not use real depths nor stereo pairs in our T$^2$Net models. The ground truth depth maps in KITTI were obtained by aligning laser scans with color images, which produced less than $5\%$ depth values and introduced sensor errors. For fair comparison with state-of-the-art single view depth estimation methods, we evaluated our results based on the cropping given in \cite{garg2016unsupervised} and clamping the predicted depth values within the range of 1--50m.

\begin{table}[tb!]
	\begin{center}
		\tiny
		\renewcommand{\arraystretch}{1.6}
		\setlength{\arrayrulewidth}{0.8pt}
		\setlength{\tabcolsep}{0.5pt}
		\caption{Results on KITTI 2015 \cite{Menze2015CVPR} using the split of Eigen \etal \cite{eigen2014depth}. For dataset, K is the real KITTI dataset \cite{Menze2015CVPR}, CS is Cityscapes\cite{Cordts2016Cityscapes} and vK is the synthetic KITTI dataset \cite{gaidon2016virtualworlds}. L, R are the left and right stereo images, and I, D are the images and depths. \emph{The gray rows highlight methods that did not use real image-depth pairs nor stereo pairs for training. Best real-supervised or stereo-based results are marked with *, while best unsupervised results are in bold.}}
		\begin{tabular}{|l|c|c|c|c|c|c|c|c|c|}
			\hline
			\multicolumn{3}{|c|}{} &  \multicolumn{4}{c|}{\cellcolor[rgb]{0.6,0.8,1.0} lower is better} &  \multicolumn{3}{c|}{\cellcolor[rgb]{0.0,0.8,1.0} higher is better} \\
			\hline
			Method & Dataset & cap & \cellcolor[rgb]{0.6,0.8,1.0} {\tiny Abs Rel} & \cellcolor[rgb]{0.6,0.8,1.0} {\tiny Sq Rel} & \cellcolor[rgb]{0.6,0.8,1.0} {\tiny RMSE} & \cellcolor[rgb]{0.6,0.8,1.0} {\tiny RMSE log} & \cellcolor[rgb]{0.0,0.8,1.0} {\tiny $\delta$<1.25} & \cellcolor[rgb]{0.0,0.8,1.0} {\tiny $\delta$<1.25$^2$}  & \cellcolor[rgb]{0.0,0.8,1.0} {\tiny $\delta$<1.25$^3$} \\
			\hline
			Eigen et al.\cite{eigen2014depth} Fine & K(I+D) & 0-80m & 0.190 & 1.515 & 7.156 & 0.270 & 0.692 & 0.899 & 0.967 \\
			\hline 
			Garg et al.\cite{garg2016unsupervised} L12 Aug.8x & K(L+R) & 1-50m & 0.169 & 1.080 & 5.104 & 0.273 & 0.740 & 0.904 & 0.962 \\  
			Godard et al. \cite{godard2017unsupervised} & CS+K(L+R) & 1-50m & 0.117 & 0.762 & 3.972 & 0.206 & 0.860 & 0.948 & 0.976 \\
			Kuznietsov et al. \cite{kuznietsov2017semi} & K(D+L+R) & 1-50m & 0.108$^*$ & 0.595$^*$ & 3.518$^*$ & 0.179 & 0.875$^*$ & 0.964$^*$ & 0.988$^*$ \\
			\hline
			\rowcolor[rgb]{0.9,0.9,0.9}
			Baseline, train set mean &  vK(I+D) & 1-50m & 0.521 & 11.024 & 10.598 & 0.473 & 0.638 & 0.755 & 0.835\\
			\hline
			Our $f_T$, all-real  & K(I+D) & 1-50m & 0.114	& 0.627& 3.549 & 0.178$^*$ & 0.867 & 0.960 & 0.986 \\
			\rowcolor[rgb]{0.9,0.9,0.9}
			Our $f_T$, all-synthetic  & vK(I+D) & 1-50m & 0.278 & 3.216 & 6.268 & 0.322 & 0.681 & 0.854 & 0.929\\
			\rowcolor[rgb]{0.9,0.9,0.9}
			Our T$^2$Net, $D_\text{feat}$ only & vK(I+D) + K(I) & 1-50m &  0.233 & 2.902 & 6.285 & 0.300 & 0.743 & 0.880 & 0.938\\
			\rowcolor[rgb]{0.9,0.9,0.9}
			Our T$^2$Net, $D_\text{image}$ only&  vK(I+D) + K(I) & 1-50m & {\bf 0.168} & {\bf 1.199} & {\bf 4.674} & {\bf 0.243} & {\bf 0.772} & {\bf 0.912} & {\bf 0.966}\\
			\rowcolor[rgb]{0.9,0.9,0.9}
			Our full T$^2$Net &vK(I+D) + K(I) & 1-50m & 0.169 & 1.230 & 4.717& 0.245 & 0.769 &    {\bf 0.912} &   0.965  \\
			\hline
		\end{tabular}
		\label{table:outdoor}
	\end{center}
\end{table}

\begin{figure}[tb!]
	\centering
	\includegraphics[width=\textwidth]{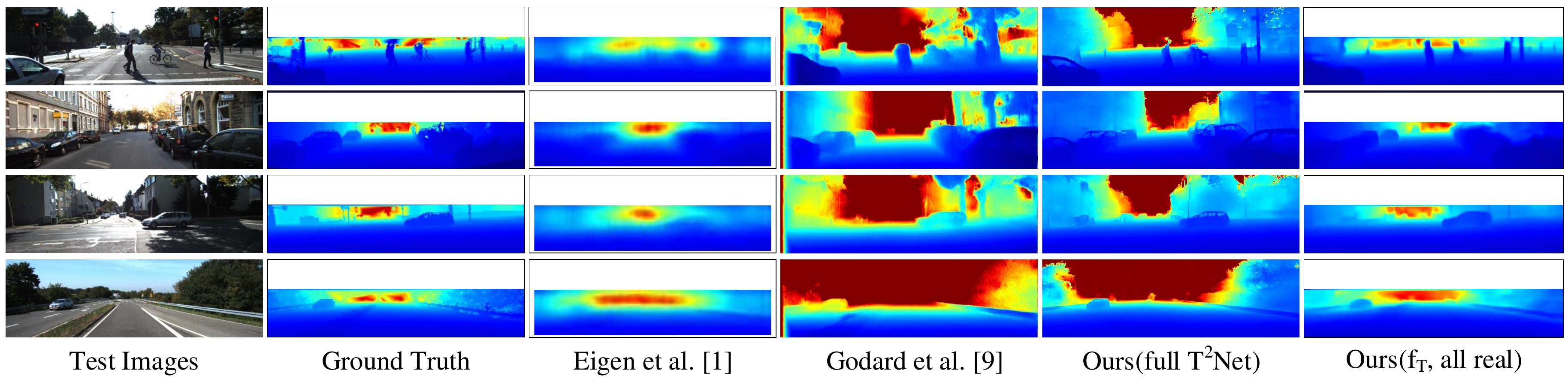}
	\caption{Qualitative results on KITTI, Eigen split \cite{eigen2014depth}. The ground truth depths in the original dataset were very sparse and have been interpolated for visualization.We converted the disparity maps provided in \cite{godard2017unsupervised} to depth maps.}
	\label{fig:depth_outdoor}
\end{figure}

Table \ref{table:outdoor} shows quantitative results of testing with real images of the KITTI dataset. We can observe that the performance of T$^2$Net has a substantial 9.1\% absolute improvement compared to our all-synthetic trained model. Unlike the indoor results, the best performance comes from without $D_{feat}$. This is likely due to the translated images much closer to real KITTI, which does not need to match the feature distribution using $D_{feat}$ adversarial learning. We also observe that our model. despite training without real paired data, is able to outperform the method of \cite{eigen2014depth} trained on real paired image-depth data, as well as the method of  \cite{garg2016unsupervised} trained on real left-right stereo data.

We also qualitatively compared the performance of the proposed model with the state-of-the-art in fig.~\ref{fig:depth_outdoor}. We only chose two representatives that either used real paired color-depth images \cite{eigen2014depth}, or real left-right stereo images \cite{godard2017unsupervised}. Compared to \cite{eigen2014depth}, our model can generate full dense depth maps of input image size. Our method is also able to detect more detail at object boundaries than \cite{godard2017unsupervised}, with a likely reason being that the synthetic training depth maps preserved object details better. Another interesting observation is the predicted depth maps were treating glass windows as permeable based on synthetic data, while they were mostly sensed as opaque in the laser-based ground truth.

\begin{figure}[tb!]
	\centering
	\includegraphics[width=\textwidth]{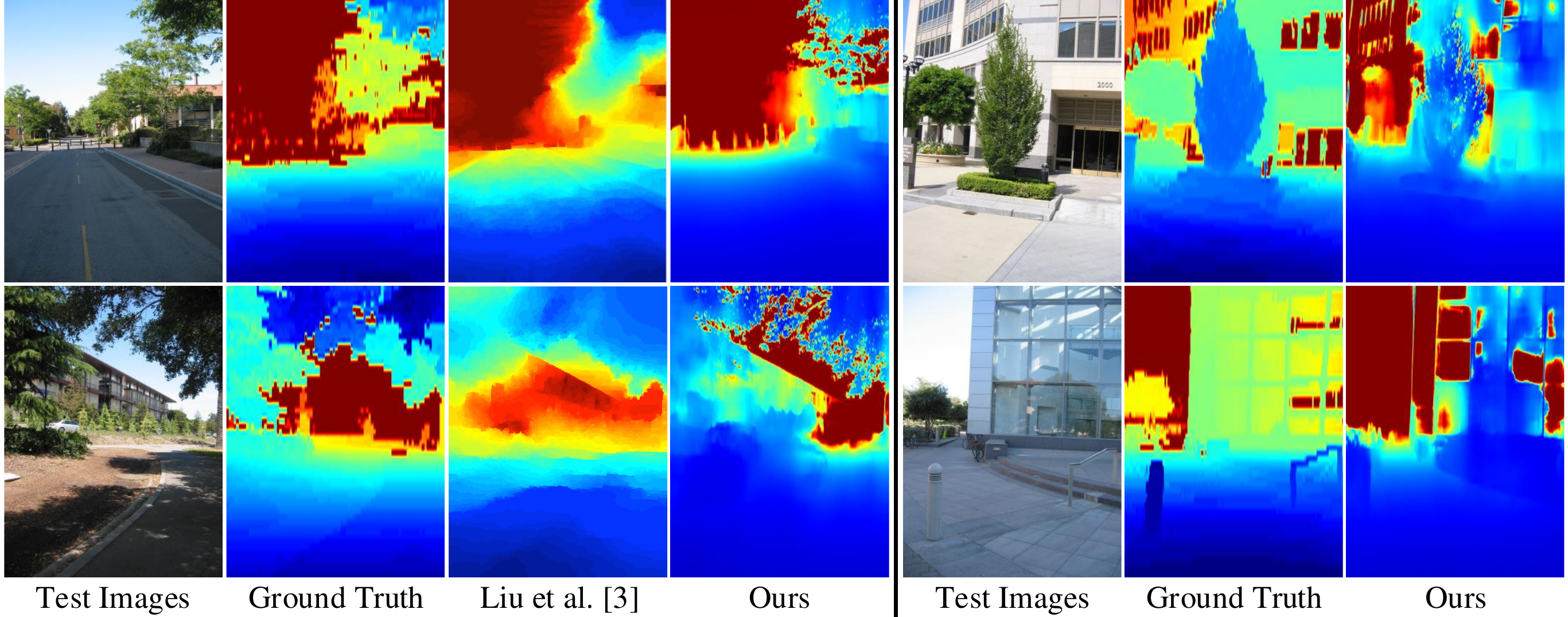}
	\caption{Qualitative results on Make3D. For most cases the model generated reasonable depths except scenes with new object types not present in the synthetic data. }
	\label{fig:depth_outdoor3D}
\end{figure}

\paragraph{\bf Performance on Make3D:} To compare the generalization ability of our T$^2$Net to a different test dataset, we used our full T$^2$Net model, trained only on vKITTI paired data and (unpaired) real KITTI images, for testing on the Make3D dataset\cite{saxena2009make3d}. We evaluated our model quantitatively on Make3D using the standard C1 metric. The RMSE(m) accuracy is 8.935, Log-10 is 0.574, Abs Rel is  0.508 and Sqr Rel is 6.589. The qualitative results presented in fig.~\ref{fig:depth_outdoor3D} show that our model can generate reasonable depth map in most situations. The right part of fig.~\ref{fig:depth_outdoor3D} displays some failure cases, likely due to large building windows not being widely observed in the vKITTI datasets.

\subsection{Ablation Study}

\begin{figure}[tb!]
	\centering
	\includegraphics[width=\textwidth]{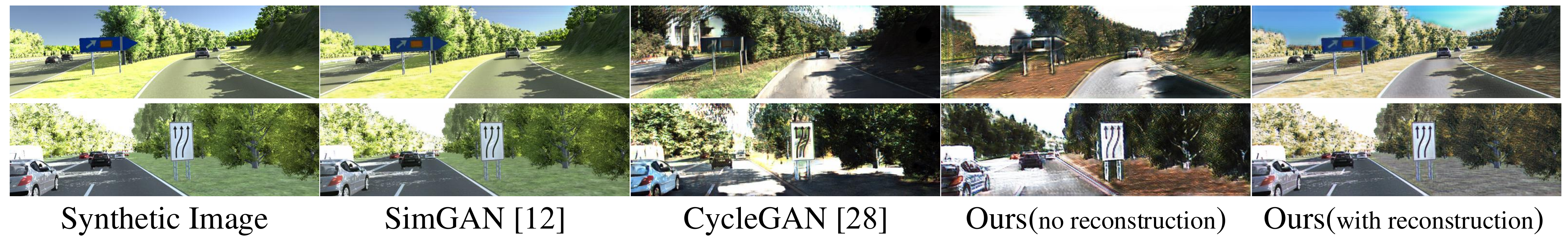}
	\caption{The qualitative results of different unpaired image-to-image translation methods trained using vKITTI and real KITTI dataset. }
	\label{fig:syn2real_ablation}
\end{figure}

We evaluated the contribution of different design choices in the proposed T$^2$Net. Table \ref{table:ablation} shows the quantitative results and fig.~\ref{fig:syn2real_ablation} shows some example outputs of different methods for unpaired image translation.

\begin{table}[tb!]
	\begin{center}
		\scriptsize
		\renewcommand{\arraystretch}{1.4}
		\setlength{\arrayrulewidth}{1.0pt}
		\caption{Quantitative results of different variants of our T$^2$Net on KITTI using the split of \cite{eigen2014depth}. All methods are trained without the real world ground truth.}
		\begin{tabular}{|l|c|c|c|c|c|c|c|}
			\hline
			\multicolumn{1}{|c|}{} &  \multicolumn{4}{c|}{\cellcolor[rgb]{0.6,0.8,1.0} lower is better} &  \multicolumn{3}{c|}{\cellcolor[rgb]{0.0,0.8,1.0} higher is better} \\
			\hline
			Method & \cellcolor[rgb]{0.6,0.8,1.0} {\tiny Abs Rel} & \cellcolor[rgb]{0.6,0.8,1.0} {\tiny Sq Rel} & \cellcolor[rgb]{0.6,0.8,1.0} {\tiny RMSE} & \cellcolor[rgb]{0.6,0.8,1.0} {\tiny RMSE log} & \cellcolor[rgb]{0.0,0.8,1.0} {\tiny $\delta$<1.25} & \cellcolor[rgb]{0.0,0.8,1.0} {\tiny $\delta$<1.25$^2$}  & \cellcolor[rgb]{0.0,0.8,1.0} {\tiny $\delta$<1.25$^3$} \\
			\hline
			baseline, synthetic only & 0.278 & 3.216 & 6.268 & 0.322 & 0.681 & 0.854 & 0.929 \\
			vanilla task network, synthetic only & 0.295 & 3.793 & 8.403 & 0.363 & 0.600 & 0.817 & 0.912 \\
			vanilla task network, full approach & 0.259 & 2.891 & 6.380 & 0.324 & 0.694 & 0.853 & 0.927\\
			\hline
			separated training & 0.234 & 2.706 & 6.068 & 0.293 & 0.747 & 0.882 & 0.942 \\
			separated training with CycleGAN & 0.212 & 1.973 &  5.340 & 0.269 & 0.750 & 0.895 & 0.952\\
			self-domain reconstruction &  0.199 & 1.517 & 5.349 & 0.298 & 0.695 & 0.866 & 0.9420 \\
			\hline
			No reconstruction loss(epoch 3)& 0.201 & 1.941 & 5.619 & 0.286 & 0.741 & 0.882 & 0.945 \\
			No feature loss & {\bf 0.168} & {\bf 1.199} & {\bf 4.674} & {\bf 0.243} & {\bf 0.772} & {\bf 0.912}& {\bf 0.966}\\
			No image GAN loss & 0.233 & 2.902 & 6.285 & 0.300 & 0.743 & 0.880 & 0.938 \\
			\hline
			our full approach & 0.169 & 1.230 & 4.717& 0.245 & 0.769 &    0.912 &   0.965 \\
			\hline
		\end{tabular}
		\label{table:ablation}
	\end{center}
\end{table}

\paragraph{\bf End-to-End vs Separated:} We began by evaluating the effect of end-to-end learning. We found that end-to-end training outperformed separated training of the translation network and task prediction network. One reasonable explanation is that task loss is a form of supervised loss for synthetic-to-realistic translation. This incentivizes the translation network to preserve geometric content present in a synthetic image.

We also experimented with the unpaired image translation network CycleGAN \cite{zhu2017unpaired}. This model has two encoder-decoder translation networks and two discriminators, but we were limited by machine memory and trained the CycleGAN and task network separately. From fig.~\ref{fig:syn2real_ablation}, we found that while this model generated very visually realistic images, it also created some realistic-looking details that significantly distorted scene geometry. The quantitative performance is close to our separated training results.

\paragraph{\bf No Image Reconstruction:} We studied what happens when training without real-image reconstruction loss. In fig.~\ref{fig:syn2real_ablation}, we may surmise that the task loss in the depth domain is able to encourage reasonable depiction of scene geometry in the translation network. However the lack of a real image reconstruction loss appears to make it harder to generate high resolution images. In addition, we noticed that while the removal of reconstruction loss still led to relatively good results as seen in table \ref{table:ablation}, this was only true in early training with best results in epoch 3, with accuracy dropping after more training epochs.

\paragraph{\bf Target Reconstruction vs Self-Regularization:} Since the self-regularization component of SimGAN is closest to our target-domain reconstruction concept, we also trained our full model with L1 reconstruction loss for synthetic imagery, which forces the generated target images to be similar to original input images. From fig.~\ref{fig:syn2real_ablation}, we observe that this is unable to work well for large domain shifts for the GAN loss and self-domain reconstruction loss play opposite roles in the translation task.  

%------------------------------------------------------------------------
\section{Conclusion and Future Work}

We presented our T$^2$Net deep neural network for single-image depth estimation, that requires only synthetic image-depth pairs and unpaired real images for training. The overall system comprises an image translation network and a depth prediction network. It is able to generate realistic images via a learning framework that combines adversarial loss for synthetic input and target-domain reconstruction loss for real input in the translation network, and a further combination of a task loss and feature GAN loss in the depth prediction network. The T$^2$Net can be trained end-to-end, and does not require real image-depth pairs nor stereo pairs for training. It is able to produce good results on the NYUDv2 and KITTI datasets despite the lack of access to real paired training data, and even outperformed early deep learning methods that were trained on real paired data. In future, we intend to explore mechanisms that provide greater generalization capability across different datasets.

\paragraph{\bf Acknowledgements} This research is supported by the BeingTogether Centre, a collaboration between Nanyang Technological University (NTU) Singapore and University of North Carolina (UNC) at Chapel Hill. The BeingTogether Centre is supported by the National Research Foundation, Prime Minister’s Office, Singapore under its International Research Centres in Singapore Funding Initiative.

\clearpage

\bibliographystyle{splncs04}
\bibliography{mybibliography}
%Author, F.: Article title. Journal \textbf{2}(5), 99--110 (2016)

%\end{thebibliography}
\end{document}